\documentclass{article}
\usepackage{ijcai23}
\usepackage{mathtools}
\usepackage{times}
\usepackage{soul}
\usepackage{url}
\usepackage[hidelinks]{hyperref}
\usepackage[utf8]{inputenc}
\usepackage[small]{caption}
\usepackage{graphicx}
\usepackage{amsmath}
\usepackage{amsthm}
\usepackage{amsfonts}
\usepackage{booktabs}
\usepackage{algorithm}
\usepackage{algorithmic}
\usepackage{multirow}

\usepackage[switch]{lineno}

\newcommand{\model}{Multiparty-Transformer}
\newcommand{\modelshort}{Multipar-T}
\newcommand{\modelss}{Multiparty-Transformer }
\newcommand{\modelshorts}{Multipar-T }

\newcommand{\cpa}{Crossperson Attention}
\newcommand{\cpas}{Crossperson Attention }
\newcommand{\cpashort}{CPA}
\newcommand{\cpashorts}{CPA }

\newcommand{\cpt}{Crossperson Transformer}
\newcommand{\cpts}{Crossperson Transformer }
\newcommand{\cptshort}{CPT}
\newcommand{\cptshorts}{CPT }

\title{Multipar-T: \underline{Multipar}ty-\underline{T}ransformer for Capturing Contingent Behaviors in Group Conversations}

\author{
Dong Won Lee%$^1$
\and
Yubin Kim \and%$^1$
Rosalind Picard \and%$^1$\and
Cynthia Breazeal \and%%$^{1}$\and
Hae Won Park %$^{1}$
\affiliations
 Massachusetts Institute of Technology\\ %$^1$
\emails
\{dongwonl, ybkim95, picard, cynthiab, haewon\}@mit.edu,
}

\begin{document}

\abovedisplayskip=-0.1cm
\abovedisplayshortskip=-0.3cm
% \belowdisplayskip=0.7cm
% \belowdisplayshortskip=0.4cm

\maketitle

\begin{abstract}

 As we move closer to real-world AI systems, AI agents must be able to deal with \emph{multiparty} (group) conversations. Recognizing and interpreting multiparty behaviors is challenging, as the system must recognize individual behavioral cues, deal with the complexity of multiple streams of data from multiple people, and recognize the subtle contingent social exchanges that take place amongst group members. To tackle this challenge, we propose the \modelss (\modelshort), a transformer model for multiparty behavior modeling. The core component of our proposed approach is the \cpa, which is specifically designed to detect contingent behavior between pairs of people. We verify the effectiveness of \modelshorts on a publicly available video-based group engagement detection benchmark, where it outperforms state-of-the-art approaches in average F-1 scores by 5.2\% and individual class F-1 scores by up to 10.0\%. Through qualitative analysis, we show that our \cpas module is able to discover contingent behavior.

\end{abstract}

\vspace{-4mm}
\section{Introduction}
% https://onlinelibrary.wiley.com/doi/full/10.1002/pits.20516

% https://journals.sagepub.com/doi/abs/10.1177/0271121413484595

% https://www.sciencedirect.com/science/article/pii/S0167639310000312

% Socially resonant behavior 

% the concept of social resonance elucidates the idea that 

 % reduce the gap between the level of communicative ability of a human and an embodied AI agent

% \textbf{Group dynamics modelling}
In order to develop AI agents that can co-exist with people in the real-world, it must be able to understand people’s behavior in a multiparty (group) setting, as many common forms of important communicative behavior take place in small group settings. Accurate recognition and interpretation of multiparty behavior enables AI agents to support and facilitate group conversations across many domains, including educational lessons, business meetings, and collaborations at the workplace. Importantly, due to COVID-19 and the proliferation of hybrid work, there is an urgent need of modeling multiparty behavior in online settings, as meetings are transitioning from in-person to online in today’s society. Hence, in this work, we focus on online multiparty interaction settings. 

% is timely and crucial

% there is a urgent need for an AI agent that can recognize and model group behavior in online settings.
% In this work, .
% Hence, an AI system capable of interacting with groups may be able to support various quintessential services that are facing shortages in the workforce, as a teaching aid, caregiver, or a   companion. 

Recognizing and interpreting group behaviors is much more challenging than that of individual behaviors. Firstly, the system must perform well in recognizing individual behavioral cues. Secondly, it must do so simultaneously, while keeping track of every individual in the group. Finally, it must also recognize the subtle interactions that take place between group members as it can provide more insights into what is being communicated. Natural human conversations are \textbf{interactively contingent}, where people act and react in a coordinated fashion in turns \cite{kopp2010social}. Consequently, understanding group behavior in multiparty conversations requires recognizing contingent behaviors between group members.

To address these challenges, we propose the \modelss (\modelshort), which is able to handle multiple streams of input data for all of the members of the group. At the core of \modelshorts is \cpa. Instead of using cross attention to discover alignment between two sequences of different modalities (i.e. vision and language)~\cite{tsai2019multimodal} or differing views of a single visual input to learn multi-scale feature representations as in previous approaches \cite{chen2021crossvit}, we propose and show that cross attention can be effectively used to capture contingent behavior between two behavioral sequences across pairs of people; we call this \cpas (\cpashort). \cpashorts implicitly searches for how and when one person's current behavior is contingent on another person's past behavior, whereas previous approaches that do not take contingent behaviors  into account. Via careful construction of the direction of attention, \cpas controls the direction of contingent behaviors it captures. Furthermore, applying multiple layers of crossperson attention allows the model to discover relationships between the contingent behaviors with the other parts of the target person's behavioral sequence. We also include a self-transformer to address cases where behaviors are non-contingent and need to rely solely on the target person's behaviors. In summary, given a target person's input behavioral sequence, \modelss applies \cpas with the behavioral sequences of 
 other members of the group in a pairwise manner to output an embedding that has contextualized information about the rich, social contingent interactions in reference to the target person's behaviors. 

 In order to measure the effectiveness of our proposed approach, we focus on the important task of engagement prediction in online learning activities. Engagement prediction is a task that requires understanding of contingent behaviors, as many studies demonstrate that the presence and lack of contingent behaviors influence people's engagement \cite{masek2021language,xu2022contingent,boyd2006contingent,sage1999peer}. Furthermore, engagement detection in group settings is an important problem in developing AI systems that can gauge a group's interest to develop behavior policies and strategies to maximize the group’s overall satisfaction with the AI agent's actions. We establish baselines to compare against the proposed model on a publicly available group engagement detection dataset in an online educational setting~\cite{reverdy2022roomreader}, where we find that \modelshorts consistently outperforms previous approaches, across all levels of engagement. We provide in-depth ablation studies to show how each specific components in \modelshorts contributes to the performance boost. Furthermore, we empirically show that the \cpas mechanism is able to discover contingent behaviors across pairs of people, which is especially important with the new EU policies \cite{european2021proposal} requiring explainability in affect recognition models. 

 % \textcolor{blue}{In summary, our paper contributes to multiparty behavior modelling in group conversations. \modelss (\modelshort) is a novel transformer model which attends to the multiple streams behavioral data in the multiparty setting. The key component of \modelshorts is \cpa, where we utilize cross attention to discover contingent human behavior. Unlike previous approaches that use cross attention to model multiview \cite{chen2021crossvit}, crossmodal\cite{tsai2019multimodal}, dyadic interactions \cite{curto2021dyadformer}, a crucial distinction is that we explicitly reverse the direction of cross attention, such that the outputting embedding prioritizes parts of target's behavioral sequence that is contingent on another person's behavior. We additionally include a self-attention module to further contextualize the embedding with the target person's own behavior and to handle non-contingent behaviors. Combined, we find that \modelss significantly outperforms all previous state-of-the-art approaches on a relevant, important and timely task of multiparty engagement detection.}

Our contributions are summarized as follows: (1) We introduce the \modelss (\modelshort), a novel transformer model which can handle multi-stream multi-party data in group conversations. (2) The key component is \cpa, which is the first of its kind to reframe cross attention to discover contingent behavior between group members by controlling the direction of the attention. The output embedding prioritizes parts of the target person $self$'s behavioral sequence that is contingent on another person $other$'s behavior. (3)  The inclusion of the self-transformer module further contextualizes the embedding with the target person's own behavior and handles non-contingent behaviors. (4)  We run extensive experiments on a important and timely task of multiparty engagement detection in online educational setting and show that \modelshorts significantly outperforms all previous state-of-the-art approaches.

\vspace{-2mm}
\section{Related Works} 
\label{sec:related}

\subsection{Contingent Behavior}
\label{sec:related_contingent_behavior}
\vspace{-1mm}
Contingent behavior refers to a person's action that takes place as a response to another person's behavior. The term falls under a broader umbrella of “inter-personal coordination{”} \cite{bernieri1988coordinated}, which refers to people's behavioral adaptations that happen as a result of social resonance in natural interaction. An example of contingent behavior is mimicry and interactional synchrony. Furthermore, nonconscious contingent behavior acts as a “social glue{”} to enhance the naturalness and sympathy in conversation \cite{lakin2003chameleon}. Temporal coordination between people in communication is found in body movements and facial expressions \cite{bernieri1994interactional}. With regards to engagement, studies have shown that contingent and reciprocal interaction amongst groups of peers \cite{sage1999peer}, teachers and students \cite{boyd2006contingent},  caregivers and children \cite{masek2021language,chen2022dyadic}, children and on-screen characters \cite{xu2022contingent}, robots and humans \cite{admoni2014data,park2017hri,chen2020teaching}, influences individual engagement, motivation, and learning. 

\vspace{-1mm}
 \subsection{Modeling of Group Interactions}
\vspace{-1mm}
In small group interactions (at least 3 people), each person has their own attributes, and each member of the group communicates with each other via both nonverbal and verbal behaviors \cite{Adamssmallgroup}. Graphical models of explicitly modeling people's interactions have been explored in various tasks such as prediction of  group performance \cite{gr_perf}, group behavior recognition \cite{gr_behav}, social interaction field modelling \cite{zhou2019social}, and social relation recognition \cite{gr_sr1}. The above-mentioned works all involve representing each person's individual features as a node, and their interactions as their edges. There has been work that utilizes attention in modeling 2-person interactions \cite{curto2021dyadformer}, however, our work addresses a more complex multiparty interaction setting with a novel set-up of cross attention which captures contingent behavior in all pairwise interactions.

% Specifically, there has been work in representing a small group as a graph for the task of group performance outcome \cite{gr_perf}, group behavior recognition \cite{gr_behav}, and social relation recognition \cite{gr_sr1, gr_sr2, gr_sr3, gr_sr4}. 

% Similar to our approach, a  cross-attention mechanism is proposed for dyadic (2-person) interactions \cite{curto2021dyadformer} to allow information flow between two people for personality prediction.

% To the best of our knowledge, our work is the first to focus on the multiparty (3+ person) setting to model all pairwise interactions between group members via cross attention to address contingent behavior.

% Similar to our approach, a  cross-attention mechanism is proposed for dyadic (2-person) interactions \cite{curto2021dyadformer} to allow information flow between two people for personality prediction. However, our work is focused on the multiparty (3+ person) setting, where our model captures all pairwise interactions between group members, and our cross-person attention mechanism is specifically designed to look for contingent behavior, grounded in psychology literature. 
% \textbf{SANDY PENTLAND - small group, amicorpus, CHI IUI}
\vspace{-1mm}
\subsection{Engagement Prediction}
\vspace{-1mm}
At a high level, engagement is defined as a state of consciousness where a person is fully immersed in the task at hand \cite{ren2016rethinking}. Studies have investigated finer differences between specific types of engagement and shown that engagement is defined to be a multi-dimensional construct, composed of \cite{fredricks2004school,silpasuwanchai2016developing} behavioral (e.g. \cite{griffin2008behavioral}), cognitive (e.g. \cite{corno1983role}), emotional (e.g. \cite{park2012makes}), and attentional (e.g \cite{chapman1997models}) engagement. In our work, we focus on perceived behavioral and emotional engagement.  Previous approaches utilize CNN-LSTM models to predict engagement \cite{del2020you,steinert2020towards}. More recently, models that use bootstrapping and ensembling are proposed in BOOT \cite{wang2019bootstrap} and ENS-MODEL \cite{thong2019engagement}. HTMIL uses a Bi-LSTM with multi-scale attention and clip-level and video-level objectives \cite{ma2021hierarchical}, and TEMMA \cite{chen2020transformer} utilizes a Resnet-Transfomer model. Unlike our work, previously proposed approaches do not take into account the group setting; they focus on modeling individuals. Closest to our work in multiparty engagement prediction is the work of \cite{zhang2022engagement}, where they utilize a graph attention network (GAT) to contextualize social interactions between multiple people to estimate engagement in elderly multiparty human-robot settings. To the best of our knowledge, we are the first to utilize a transformer network to model group's behavioral contingencies for engagement prediction.

\vspace{-2mm}
\section{Problem Statement}
\vspace{-1mm}
\label{sec:ps}
\label{sec:problem}
 
 We formulate the multiparty video-based engagement prediction problem as the following. We are given video clips of groups involved in an online learning activity. We split the videos into N interval clips of $k$ frames. At any arbitrary time $t$, where $t$ is the exact timestep in which we want to predict each individual's engagement value, we are given the $[t-k,...,t]$ interval of contextual video information; $k$ is the number of frames we will use as context. Let $P$ be the number of all participants in the video, for a person in the clip $p \in [P]$, their corresponding contextual behavioral features can be viewed as  $X^t_p = [x^{t - k}_p, \ldots, x^{t}_p]$, where $x^t_p \in \mathcal{R^{F}}$ with dimension size $F$ at the $t^{th}$ frame. For brevity, we will drop the $t$ and assume it is arbitrarily fixed. The input with all of the group's features is a 3-D tensor, $X = [x_1, \ldots, x_P] \in \mathcal{R}^{ P \times k \times F}$. For a target person: $self$, we train a model that takes as input $X$, which includes the target person's features $x_{self}$ as well as all other members' features $[x_{other}$, $\forall other \in P \setminus \{ self \} ]$ to predict the engagment value $\hat{Y}_{self} \in (0,1)^C$. 
 
 % \textbf{add self, other here?}

%  $[x_1, \ldots, x_P ]$ we want to train a model  and  , where $C$ corresponds to a probability distribution over $C$ discrete engagement classes. We are training a function $f$: 

%   \begin{align}
%             \hat{Y_p} =  f(X_p, X^2_t, \ldot, X^2_, \lambda)
% \end{align}

%  a matrix of size $Y_t\in P \times C$ 

\begin{figure*}[t]
    \vspace{-0mm}
    \begin{center}
    \hspace*{-0.3in}
    \includegraphics[width=1.1\textwidth]{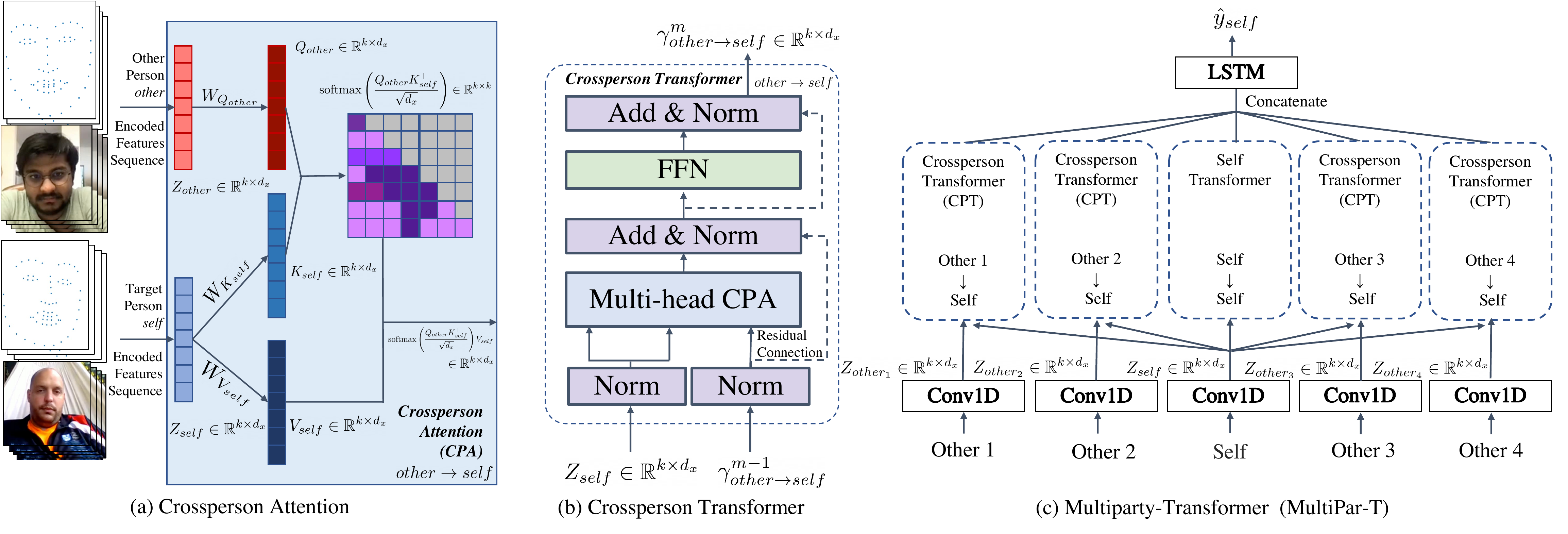}
    \caption{Full model architectures(a) Diagram of our \cpas ($\text{CPA}_{other \rightarrow self}$) module, which  automatically searches for $self$'s behaviors that are contingent on $other$'s behaviors. (b) Diagram of the proposed new \cpt. (c) Diagram of the overarching new model: \model  \ takes in all other persons's behavioral features,  applies \cpas w.r.t to $self$'s features in the \cpt, as well as self-attention in the Self Transformer. Best viewed zoomed in and in color.}
    \vspace{-6mm}
    \label{fig:fullmodel}
    \end{center}
\end{figure*}

\vspace{-2mm}
\section{Methods}
Here, we describe our proposed \underline{\textbf{Multipar}}ty-\underline{\textbf{T}}ransformer (\textbf{\modelshort}). We utilize the \cpas (\cpashort) module to discover the contingencies across time-series sequences of behaviors from pairs of people in the group.

% We perform \cpashorts across all pairs of members in the group, to capture how each person's behavior is affected by all other members' behavior in the group. The resulting group representation is contextualized with contingent behaviors.

% \ , which contextualizes the embedding with information about contingent behavior amongst participants in the group interaction.

% , which is fed into a recurrent model to enforce a greater temporal structure and predict the final output. 

\vspace{-2mm}
\subsection{\cpa}
\label{subsec:cpa}

Given a pair of people, we want to capture their contingent behaviors. We state that target person $self$'s behavior is \emph{contingent on} person $other$'s behavior if person $self$'s behavior was likely to be influenced by person $other$'s behavior ($other$$\rightarrow$$self$). We posit that capturing contingent behavior across people can be handled with mechanisms that can capture alignment between sequences. Inspired by the multimodal transformer model \cite{tsai2019multimodal}, which shows one effective method that can automatically align sequences of differing modalities is by using a scaled dot product cross attention \cite{chen2021crossvit,vaswani2017attention}, we propose that it could be applied to pairs of behavioral features, which we call \cpa \text{ } (\cpashort), to automatically discover contingent behavior and subtle social interactions. 

Cross attention utilizes query $Q$, key $K$, and values $V$, where the first step is to find the importance of each key with respect to the query. The attention mechanism computes the dot product of the query with each key to obtain a weight for each key. These resulting weights represent the importance of each key to the query. The weights are then used to obtain a weighted sum of the values in the matrix, which is referred to as the context vector. Therefore, for the target person $self$, and another person $other$, we are given their time-aligned encoded visual representations $Z_{self}$, $Z_{other} \in$ $\mathbb{R}^{k \times d_x}$, where $d_x$ is the dimension size of the visual embeddings. Therefore, we construct the queries, keys, and values as the following:  $\mathrm{Query}$s as $Q_{other} = Z_{other} W_{Q_{other}}$, $\mathrm{Key}$s as $K_{self} = Z_{self} W_{K_{self}}$, and $\mathrm{Value}$s as $V_{self} = Z_{self} W_{V_{self}}$, where $W_{Q_{other}}, {W_{K_{self}}}, W_{{V_{self}}} \in \mathbb{R}^{d_x \times d_x}$ are trainable weight parameters. 

% (i.e. which part of self behavior is related or triggered by another person's behavior)

% \begin{equation}
% \label{eq:crosspersonattn}
% \resizebox{0.9\hsize}{!}{
% $
% \begin{multlined}
%     \text{CPA}_{other \rightarrow self}(Z_{other}, Z_{self})
%     = \text{softmax}\left(\frac{Q_{other} K_{self}^\top}{\sqrt{d_x}}\right) V_{self} \\
%     &= \text{softmax}\left(\frac{Z_{other} W_{Q_{other}} \left(Z_{self} W_{K_{self}}\right)^\top}{\sqrt{d_x}}\right) Z_{self} W_{V_{self}}.
% \end{multlined}
% $
% } 
% \end{equation}

\begin{equation}
\begin{aligned}
& \textstyle{\mathrm{CPA}_{other \rightarrow \text self}\left(Z_{other}, Z_{self}\right)=\operatorname{softmax}\left(\frac{Q_{other} K_{self}^{\top}}{\sqrt{d_x}}\right) V_{self}} \\
& =\operatorname{softmax}\left(\frac{Z_{other} W_{Q_{other}}\left(Z_{self} W_{K_{self}}\right)^{\top}}{\sqrt{d_x}}\right) Z_{self} W_{V_{self}} .
\end{aligned}
\end{equation}

We refer the readers to Figure \ref{fig:fullmodel}(a) for a visual depiction. In Equation \ref{eq:crosspersonattn}, the scaled softmax produces the attention weight between two people's temporal behavior inputs, which weighs the importance of person $other$'s each behavioral timesteps with respect to the $self$'s behavior. Specifically, the resulting weight is a $k \times k$ matrix, where $k$ is the number of timesteps in the sequence. After the dot poduct with $V_{self}$, the \cpas from $other$ to $self$  $\text{CPA}_{other \rightarrow self}(Z_{other}, Z_{self})$ outputs an embedding which has captured the person $self$'s behavior contingent on person $other$'s behaviors. We highlight this is the \emph{reverse} direction of cross attention compared to many previous works \cite{tsai2019multimodal,curto2021dyadformer}, and a crucial distinction in capturing contingent behaviors as we discuss in Section \ref{ssec:directions}. Furthermore, \cpas mechanism is performed with $h$ multiple heads; we define this as $\text{CPA}^{multi}_{other \rightarrow self}$.

% Each entry captures the attention of the pair's behavior at that timestep.

% Consequently, the output of \cpashort, which is the product between the attention weights and behavioral features $V_{self}$, contains information about person $self$'s behavior that is contingent on the $other$'s behavior. 

% The $\left(i, j\right)$-th entry represents the attention between the behavior present in the $i$-th frame in person $self$ with the behavior in the $j$-th frme in person $other$. 

\begin{equation}
\label{eq:multihead}
\begin{aligned}
& \operatorname{CPA}_{other \rightarrow self}^{multi}\left(Z_{other}, Z_{self}\right) \\
& =\operatorname{Concat}\left(\mathrm{CPA}_{other \rightarrow self}^1, \ldots \operatorname{CPA}_{other \rightarrow self}^h\right) W^{\text {multi }}
\end{aligned}
\end{equation}

% \begin{equation}
% \begin{align}
%     & \text{CPA}^{multi}_{other \rightarrow self}(Z_{other}, Z_{self})  &4\\ 
%       &= \text{Concat}(\text{CPA}^{1}_{other \rightarrow self}, \ldots \text{CPA}^{h}_{other \rightarrow self})W^{multi}
% \end{align}
% \end{equation}

The outputs of each head of \cpashorts are concatenated, then linearly projected with weight matrix: $W^{multi} \in \mathbb{R}^{ (h \cdot d_x) \times d_x} $.

\vspace{-2mm}
\subsection{Multiparty Transformer}
\label{subsec:overall}

In order to successfully address the complex social interactions taking place in a group setting, we must properly represent each person's individual temporal features, address the group social interactions, then take into account the group's temporal nature. We describe in detail the individual components which are designed to tackle these challenges.   

\paragraph{Individual Temporal Encoder: Convolutions and Positional Encoding} 

We utilize 1D convolutional layers such that the convolution kernel convolves over the temporal dimension and each timestep in the sequence is contextualized by its surroundings. Furthermore, we further enforce the temporal structure by including the additive positional encoding (PE) used in \cite{vaswani2017attention}. Therefore, the individual temporal encoder, given target person $self$'s input $X_{self}$ is:
\vspace{-1mm}

  \begin{align}
  \label{eq:indiveq} 
    Z_{p} &= \text{Conv1D}(X_{p}) + \text{PE}(X_{p}) 
\end{align}%

 Conv1D includes a kernel that maps each individual's features into a common dimension $d_x$.

\paragraph{Behavior Interaction Encoder: \cpts \& Self Transformer}

\cpas (\cpashort) is a core component of the $M$-layered \cpts (\cptshort). $\text{CPA}_{other \rightarrow self}^{m, \text{multi}}$ refers to the multi-headed \cpas from person $other$ to person $self$ at the $m$-th layer.  Following standard transformer operations \cite{vaswani2017attention}, $\hat{\gamma}^{m}_{other \rightarrow self}$ refers to the intermediate output after the \cpas with residual connections. $\gamma^{m}_{other \rightarrow self}$ refers to the final output of a cross-person transformer block after feedforward network (FFN) and residual connections. As the input to the first layer, $\gamma^{0}_{other \rightarrow self} = Z_{other}$. We refer the readers to Figure \ref{fig:fullmodel}(b) for details.

\begin{equation}
\resizebox{0.95\hsize}{!}{
$
\begin{split} 
    \gamma^{m}_{other \rightarrow self} &= \text{CPT}^{m}_{other \rightarrow self}(\gamma^{m-1}_{other \rightarrow self}, Z_{self}) \\
    \hat{\gamma}^{m}_{other \rightarrow self} &= \text{CPA}_{other \rightarrow self}^{m, multi}(\text{Norm}( \gamma^{m-1}_{other \rightarrow self}), \text{Norm}( Z_{self}))
    \\& + \text{Norm}( \gamma^{m-1}_{other \rightarrow self})) \\
    \gamma^{m}_{other \rightarrow self} &= \text{Norm}(\text{FFN}(\hat{\gamma}^{m}_{other \rightarrow self}) + \hat{\gamma}^{m}_{other \rightarrow self})
\end{split}
$}
\end{equation}

 With this formulation, $\text{CPA}^{0}_{other \rightarrow self}(Z_{other}, Z_{self})$ discovers contingent behaviors in the first layer. Then, in the later \cptshorts layers, \cpashorts contextualizes the embedding by discovering correlations on how the contingent behavior is related to different parts of the target person's behaviors. We empirically show that standalone first layer \cptshorts is not enough, and that further contextualization with multiple layers of transformer blocks is useful in Section \ref{subsec:hyperparam}. 
 
 % importance of the multiple layers of the transformer blocks  \textbf{This is important - just the first layer is not enough -- shown with ablation studies in the hyperparams section}

In addition to all pair-wise \cpas across all other members of the group $\forall other \in P \setminus \{ self \}$, $\text{CPA}^{multi}_{other \rightarrow self}$, we also compute self-attention in order to (1) account for how one's earlier behavior correlates with their current behavior and (2) handle cases where there are no contingent behavior information. This is equivalent to performing \cpashorts with an equivalent query, key and value matrices (i.e. given a target person $self$ we perform $\text{CPA}^{multi}_{self \rightarrow self}(Z_{self}, Z_{self})$). Its usefulness is tested with ablation studies in Section \ref{subsec:hyperparam}.

\paragraph{Temporal Classifier}
Finally, we concatenate the outputs from the above-mentioned behavior interaction encoder, $[\cdot||\cdot]$ refers to concatenation.  The concatenated outputs are fed into an LSTM for $k$ steps to enforce a stronger temporal structure. The resulting output is passed through fully connected layers (FFN) for the final prediction $\hat{Y}_{self}$,

\begin{equation}
\resizebox{\hsize}{!}{
$
\begin{split}
    \zeta^{n}_{self}, hidden^{n} &= \text{LSTM}([\gamma^{M}_{1 \rightarrow self}  ||  \ldots || \gamma^{M}_{P \rightarrow self}] , hidden^{n-1} ) \\
    \hat{Y}_{self} &= \text{FFN}(\zeta^{k}_{self}) \qquad\qquad\qquad for \text{ } n \in [k]
\end{split} 
$}
\end{equation}
\vspace{-8mm}

 % We repeat the process to compute each individual's engagement predictions $\hat{Y_t} = [\hat{Y}_1, \ldots , \hat{Y}_P ]$ where $P$ is the number of people in the group.

% Formally, as input, we are given a context matrix of $\mathbf{X}$ which consists of visual features, 

% - Short description of the type of engagement, (on-task, on robot)?

% \subsection{Group Social Interactions}

% \emph{Attending over each person's behavior: }

% \emph{Data Augmentation by Shuffling Groups as negatives and Contrastive Learning: }

% \subsection{Contingent Behavior}

% \subsection{Context and Teacher Forcing}

\vspace{-2mm}

\section{Experiments}

% \usepackage{booktabs}
\begin{table*}[t]
\centering
\resizebox{1\textwidth}{!}{%
\begin{tabular}{@{}l|lll|l|l|l|l@{}}
\toprule
             & \multicolumn{3}{l|}{All Engagement Classes}    & High Dis-Eng. & Low Dis-Eng. & Low Eng.     & High Eng.    \\ \midrule
Model        & Accuracy       & Weighted F1   & Macro F1      & F1            & F1           & F1           & F1           \\ \midrule
ConvLSTM \cite{del2020you}         & 0.859 $\pm$ 0.01   & 0.857 $\pm$ 0.02  & 0.699 $\pm$ 0.05   & 0.741         & 0.459 $\pm$ 0.22 & 0.699 $\pm$ 0.12 & 0.907 $\pm$ 0.01 \\
OCtCNN-LSTM \cite{steinert2020towards} & 0.769 $\pm$ 0.08   & 0.695 $\pm$ 0.14  & 0.410 $\pm$ 0.10    & 0.588         & 0.119 $\pm$ 0.17 & 0.233 $\pm$ 0.33 & 0.864 $\pm$ 0.05 \\
TEMMA \cite{chen2020transformer}      & 0.823 $\pm$ 0.02   & 0.822 $\pm$ 0.02  & 0.561 $\pm$ 0.11  & 0.286         & 0.254 $\pm$ 0.19 & 0.621 $\pm$ 0.13 & 0.885 $\pm$ 0.01 \\
EnsModel \cite{thong2019engagement}    & 0.760 $\pm$ 0.07  & 0.675 $\pm$ 0.12 & 0.302 $\pm$ 0.03  & 0             & 0.000 $\pm$ 0.00            & 0.160 $\pm$ 0.23  & 0.860 $\pm$ 0.05  \\
BOOT  \cite{wang2019bootstrap}       & 0.817 $\pm$ 0.03  & 0.822 $\pm$ 0.03  & 0.636 $\pm$ 0.09  & 0.714         & 0.320 $\pm$ 0.24  & 0.658 $\pm$ 0.12 & 0.873 $\pm$ 0.02 \\
HTMIL  \cite{ma2021hierarchical}      & 0.820 $\pm$ 0.02  & 0.818 $\pm$ 0.02 & 0.460 $\pm$ 0.05   & 0             & 0.000 $\pm$ 0.00            & 0.633 $\pm$ 0.12 & 0.880 $\pm$ 0.02  \\

GAT \cite{zhang2022engagement}         & 0.739 $\pm$ 0.06 & 0.631 $\pm$ 0.08 & 0.261 $\pm$ 0.03 & 0             & 0.000 $\pm$ 0.00            & 0.006 $\pm$ 0.01 & 0.848 $\pm$ 0.04 \\

MulT \cite{tsai2019multimodal}    & 0.847 $\pm$ 0.02        & 0.845 $\pm$ 0.02          & 0.624 $\pm$ 0.12 & 0.625          & 0.310 $\pm$ 0.25           & 0.665 $\pm$ 0.12          & 0.901 $\pm$ 0.01 \\ \midrule

\modelshorts (Ours)         & \textbf{0.888 $\pm$ 0.03}  & \textbf{0.887 $\pm$ 0.03}  & \textbf{0.751 $\pm$ 0.05}                                    & \textbf{0.800}          & \textbf{0.559 $\pm$ 0.07}           & \textbf{0.759 $\pm$ 0.11} & \textbf{0.927 $\pm$ 0.02} \\ \bottomrule
\end{tabular}
}

\vspace{-2mm}
\caption{Results and standard deviations for engagement recognition models for 3 seeds (std dev for High Dis-Eng.  not reported due to 2 seeds not having corresponding labels). Despite high accuracy and weighted-F1 scores, many previous baselines fail at infrequent disengagement classes. \modelshorts outperforms other approaches across all metrics. }
\vspace{-2mm}
\label{tab:full_res}
\end{table*}

\subsection{Dataset}

 We utilize the RoomReader \cite{reverdy2022roomreader} as a benchmark to measure the performance of our proposed method against other baselines. RoomReader \cite{reverdy2022roomreader} is a corpus of multimodal, multiparty conversational interactions in which participants followed a collaborative online student-tutor scenario designed to elicit spontaneous speech. Engagement is focused on off-task/on-task engagement, where the task at hand is led by the instructor. 
 
 % The annotation scheme is defined to capture perceived behavioral and emotional engagement, as annotators are instructed to recognize behavioral and affective cues such as signs of intention to take the floor, walking around, and neutral expressions. 

\begin{figure}[htb]
    \begin{center}
    \includegraphics[width=1\columnwidth]{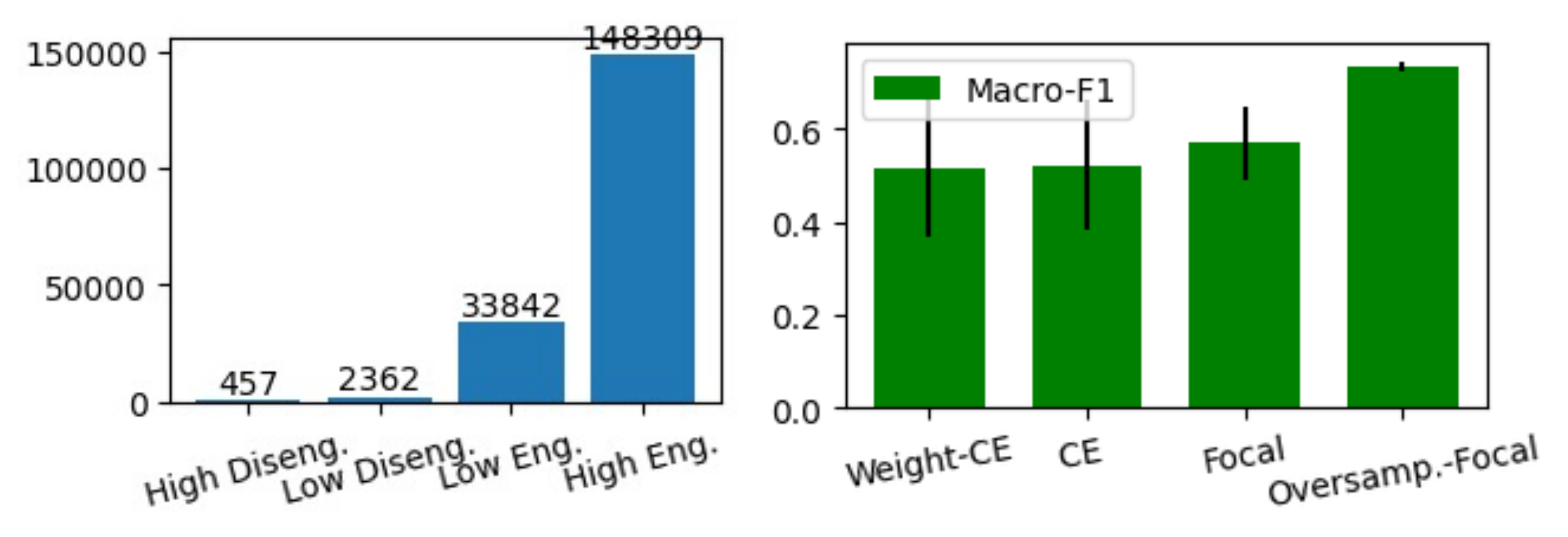}
    \caption{(Left) Distribution of class labels in the dataset. There is a severe class imbalance. (Right) Macro-F1 scores for \modelshorts for each class imbalance strategy, (CE refers to Cross-Entropy), where the combination of oversampling and focal loss performs well.}
    \vspace{-5mm}
    \label{fig:imbalanced}
    
    \end{center}
\end{figure}

\paragraph{Engagement Classification} RoomReader provides continuous annotations for engagement, where the engagement labels range from $[-2, 2]$. Instead of regression, we define the task as a 4-class classification, where labels between $(1,2]$ refer to high engagement, $(0,1]$: low engagement, $(-1,0]$: low disengagement, $(-2,-1]$: high disengagement. Setting up the task in this way results in more interpretable evaluation metrics than regression losses (such as MAE, or MSE), and allows us to report categorical metrics conditioned on each class. Generally, it is well known that class imbalance is often severe for datasets with engagement labels \cite{del2020you,dhall2020emotiw,steinert2020towards}. In the RoomReader dataset, 80.2\% of the entire dataset consists of highly engaged samples, 18.3\%: low engagement, 1.3\%: low disengagement, and 0.2\%: high disengagement. We refer the readers to Figure \ref{fig:imbalanced} for the imbalanced distribution of labels. To counter the effects of class imbalance, we (1) oversample the infrequent class to balance the dataset and (2) train the models with a Focal Loss \cite{lin2017focal} that applies a modulating term to the cross entropy loss in order to focus learning on hard misclassified examples as shown below.  $y_{i c}$ refers to the ground truth labels, $\hat{Y}_{i c}$ is the probability prediction, and $\alpha$ is a hyperparameter that weighs how much easy samples should be down-weighted. 

\begin{align}
            L_{Focal} =  -\frac{1}{N} \sum_i^N \sum_c^C (1-\hat{Y}_{i c})^{\alpha} y_{i c} \log \left(\hat{Y}_{i c}\right)
\end{align}%

Combined, we find that the models are able to predict the infrequent classes and overcome the class imbalance problem, which can be seen in Figure \ref{fig:imbalanced}.

\paragraph{Data Preprocessing} For the input features, we utilize the normalized eye gaze direction, location of the head, location of 3D landmarks, and facial action units extracted via OpenFace \cite{8373812}. In addition, we extract frame-wise image features from the penultimate layer of Resnet-50 \cite{he2016deep}. The two features are concatenated per timestep to be used as input. The input feature dimension size per timestep is $F = 2183$. For each label at timestep $t$, we use 8 seconds worth of video context information, where the frame rate is 8 fps. We utilize $k = 64$ frames as input. We apply a sliding window with an interval of 1 second between each sample. In total, we have 184970 samples. 

% \textbf{include total number of samples, split size -- just add figure 2}

 \vspace{-1mm}
\subsection{Baseline Models}
 \vspace{-1mm}
% Both study:
% Video Data
% Human Behaviors
% Similar Spatio-Temporal Reasoning

% But... 
% More technical developments for action recognition
% Unified benchmark and testing – systemic, incremental development 

% Hence:
% We develop a unified benchmark on publicly available dataset
% Release additional dataset for public use 
% We establish all of these baselines for accurate performance comparison 
% Reduce the research gap between these two equally important research problems 

We compare our proposed model with a family of baselines in engagement prediction, as well as action recognition. We run the newest versions of these models and report their scores on a unified benchmark.  We compare \modelshorts to ConvLSTM \cite{del2020you}, OCtCNN-LSTM \cite{steinert2020towards}, TEMMA  \cite{chen2020transformer}, BOOT \cite{wang2019bootstrap} and ENS-MODEL \cite{thong2019engagement}, GAT \cite{zhang2022engagement}, and MulT \cite{tsai2019multimodal}. For action recognition models, we compare our method with TimeSformer \cite{bertasius2021spacetime}, SlowFast \cite{feichtenhofer2019slowfast} and  I3D \cite{NonLocal2018}. 

% Although both require spatio-temporal understanding of human behavior data in videos, there is a large gap between these two tasks. The models that are considered state-of-the-art in engagement detection are methods that have been followed by newer methods in action recognition. To reduce the research gap between these two classes of problems, 

% , which utilize facial, skeletal key points and image features to regress to engagement labels H-TMIL uses a Bi-LSTM with a clip level and video level objective \cite{ma2021hierarchical}. TEMMA  \cite{chen2020transformer} utilizes a Resnet-Transfomer model. Unlike our work, previously proposed approaches do not take into account the group setting; they focus on modeling individuals. Closest to our work in multiparty engagement prediction is the work \cite{zhang2022engagement}, where they utilize a graph attention network (GAT) to contextualize social interactions between multiple people to estimate engagement in elderly multiparty human-robot settings. To the best of our knowledge, our work is the first to utilize the attention mechanism to capture contingent multiparty behavior.

 \vspace{-2mm}
\subsection{Implementation Details}

 We train our models on 2 NVIDIA GeForce GTX 1080 Ti with a batch size of 64 for 20 epochs. We use the AdamW \cite{loshchilov2017decoupled} optimizer with an initial learning rate of 0.0001 with a scheduler that decays the learning rate by 0.1 every 5 epochs. We train on 16 groups' data, validate on 3 groups, and test on 1 group for 3 seeds. The model is exposed to different totally held-out subsets of groups for cross-validation. Our code can be found in the Supplementary, and will be shared on a public github repository with camera ready. \modelshorts can be used in real-time, where the inference time only takes $0.0981 \pm 0.0029$ seconds.

% 98.08798184712728 millisec
% 2.9101706445858024 millisec

% \textbf{real-time performance - compute time - inference}

% \subsection{Evaluation Metrics}

 \vspace{-2mm}

% \usepackage{booktabs}
% \usepackage{multirow}
\begin{table*}[]
\resizebox{1\linewidth}{!}{%
\begin{tabular}{@{}l|l|lll|llll@{}}
\toprule
\multirow{2}{*}{Attention Direction}             & \multirow{2}{*}{Ablation}      & \multicolumn{3}{l|}{All Classes}                                                                     & High Dis-Eng.  & Low Dis-Eng.          & Low Eng.              & High Eng.             \\ \cmidrule(l){3-9} 
                                  &                                & Accuracy              & Weighted F1            & Macro F1                                                & Binary F1      & Binary F1             & Binary F1             & Binary F1             \\ \midrule
                                  & \modelshorts $w/o$ \cpt & 0.847 + 0.0154        & 0.844 + 0.14           & 0.661 + 0.018                                           & 0.588          & 0.433 + 0.1           & 0.66 + 0.12           & 0.901 + 0.01          \\ \midrule
\multirow{2}{*}{$CPA_{self \rightarrow other}$} & \modelshorts $w/o$ Self Transformer        & 0.847 + 0.0167        & 0.845 + 0.021          & 0.624 + 0.12 & 0.625          & 0.31 + 0.25           & 0.665 + 0.12          & 0.901 + 0.01          \\
                                  & \modelshorts                           & 0.865 + 0.03          & 0.862 + 0.036          & 0.735 + 0.02                                            & \textbf{0.769} & \textbf{0.587 + 0.12} & 0.698 + 0.15          & 0.912 + 0.02          \\ \midrule
\multirow{2}{*}{$CPA_{other \rightarrow self}$} & \modelshorts $w/o$ Self Transformer        & \textbf{0.883 + 0.02} & \textbf{0.884 + 0.024} & \textbf{0.75 + 0.04}                                    & \textbf{0.769} & \textbf{0.555 + 0.11} & \textbf{0.762 + 0.08} & \textbf{0.923 + 0.02} \\
                                  & \modelshorts                           & \textbf{0.883 + 0.02} & \textbf{0.885 + 0.02}  & \textbf{0.75 + 0.06}                                    & 0.714          & 0.557 + 0.19          & \textbf{0.766 + 0.08} & \textbf{0.923 + 0.02} \\ \bottomrule
\end{tabular}
}

\vspace{-3mm}
\caption{Ablation results for Self Tranformer and \cpts mechanisms. Attending to $other$'s and own $self$ behaviors boosts performance. We refer the readers to Figure \ref{fig:fullmodel}. \modelshorts $w/o$ \cpts refers to the ablation of all pairwise Crossperson Transformers with only the Self Transformer remaining. \modelshorts $w/o$ Self Transformer refers the ablation of the Self Transformer and utilizing the pairwise Crossperson Transformers. Results with different directions of \cpas are displayed, where $CPA_{other \rightarrow self}$ performs well generally and $CPA_{self \rightarrow other}$ performs well for disengaged instances. } 
\vspace{-4mm}
\label{tab:attn}

\end{table*}
\section{Results \& Discussion }

In this section, we discuss the quantitative and qualitative results of our experiments. We compare our approach \modelshorts with state-of-the-art baselines. Then, we discuss the importance of the attention modules and the effects of their directions. Finally, we qualitatively demonstrate that \cpas has learned to recognize contingent behaviors.

\subsection{Quantitative Results}

Following previous works \cite{del2020you,dhall2020emotiw,steinert2020towards}, we report accuracy and weighted-F1, which is the weighted mean of all per-class F1 scores considering each class's support in the data. Most importantly, we report the macro-F1, i.e., the unweighted mean of per-class F1. A high macro-F1 score demonstrates that the model performs well across all engagement classes regardless of its frequency in the dataset. 

\paragraph{Comparisons Against State-of-the-Art}
In Table \ref{tab:full_res}, state-of-the-art engagement prediction models are compared to \modelshort. \modelshorts outperforms all previous state-of-the-art approaches in accuracy by 2.9\%, weighted F1 by 3.0\%, and Macro-F1 by 5.2\%. Again, Macro-F1 is the most informative metric, as predicting the most infrequent class, i.e., disengagement, is the most challenging component of this problem. For a closer look into how \modelshorts performs for individual levels of engagement, we refer the readers to the right of Table \ref{tab:full_res}, where we compare each model's performance for each specific engagement class. Although other baselines result in comparable high accuracy and weighted-F1 scores, they fail to predict the infrequent disengagement class.  \modelshorts has significant performance gains (10\% increase against best performing baseline) in the most challenging task of accurately predicting high and low dis-engagement, which consists of only 2\% of the entire dataset. Moreover, comparing with the other group behavior encoding method, GAT \cite{zhang2022engagement}, we find that \modelshorts outperforms across all metrics, highlighting that our method is a more effective method of capturing group behavior. We also compare with an adaptation of the multimodal transformer (MulT) \cite{tsai2019multimodal} where we replace differing modality inputs with differing person's behavioral sequences, which is equivalent to \modelshorts $w/o$ Self Transformer and reversed attention direction $CPA_{self \rightarrow other}$ in Table \ref{tab:attn}. The inclusion of the self-transformer and the configuration of the attention directions is a key component in modeling human multiparty behavior, different from multi-modality alignment, as we demonstrate in ablation studies in the next sections. 

% ,
% performs especially well against other baselines
% as accurately predicting rare classes (disengagement labels) is equally as important as predicting common classes.

\paragraph{Importance of \cpas (\cpashort) and Self-Attention} In Table \ref{tab:attn}, we present results where we ablate \cpas and Self-Attention from our models. We see that ablating \cpas leads to a significant drop in performance metrics, especially Macro-F1. The model struggles at harder, low-data disengagement instances. Therefore, the inclusion of \cpa, which allows the model to attend to how others are behaving in the group provides more context information for the model to differentiate harder cases. We also find that the ablation of self-attention leads to significant drops in performance metrics, as self-attention provides more information regarding one's own behavior, which is especially important when there are no contingent behaviors.  

\paragraph{Importance of the Direction of CPA}
\label{ssec:directions}
In predicting a target person $self$'s engagement value, we hypothesized that the $self$'s behavior contingent on  $other$'s behavior is an important predictor of engagement and disengagement. On the other hand, we hypothesized that the $other$'s behavior contingent on the $self$'s behavior would not be an important predictor. To test these hypotheses, we carefully experiment with the directions of the \cpas mechanism. In Table \ref{tab:attn}, $\text{CPA}_{other \rightarrow self}$ refers to our set up of the attention direction, performing \cpas where the query corresponds to behaviors of \emph{other} persons in the group, and the key and value correspond to the behavior of the target, \emph{self}. The resulting embedding contains information about the $self$'s behavior which is contingent on  $other$'s. Conversely, $\text{CPA}_{self \rightarrow other}$, outputs an embedding with the $other$'s behavior contingent on the $self$'s behavior. This is similar to the cross attention set-up in \cite{tsai2019multimodal,curto2021dyadformer}.

% This contextualizes the embedding such that it contains information about how $other$ (another) person's behavior in the group is affecting the $self$ or Person $self$'s behaviors.

% The experiments with different directions of cross person attention shed interesting insights regarding the problem of engagement detection. 

We refer the readers to the results for \modelshorts in $\text{CPA}_{other \rightarrow self}$ and $\text{CPA}_{self \rightarrow other}$. We find that \modelshorts with our formulation of cross attention, $\text{CPA}_{other \rightarrow self}$, performs significantly better, which indicates that it is important to explicitly set the direction of cross attention such that the output embedding prioritizes parts of target's behavioral sequence that is contingent on another person's behavior. Interestingly, when predicting low disengagement, \modelshorts with $\text{CPA}_{self \rightarrow other}$ results in better performance. This shows that how the $self$ impacts $others$ could be a important predictor when predicting if the target person is disengaged. 

% (i.e. $self$ is disengaged and starts fidgeting, $other$ person notices and starts stares at $self$ -- noticing that $other$'s behavior is contingent on $self$'s behavior could be an important predictor for predicting disengagement). 

\begin{figure}[t]
    \begin{center}
    \includegraphics[width=1\columnwidth]{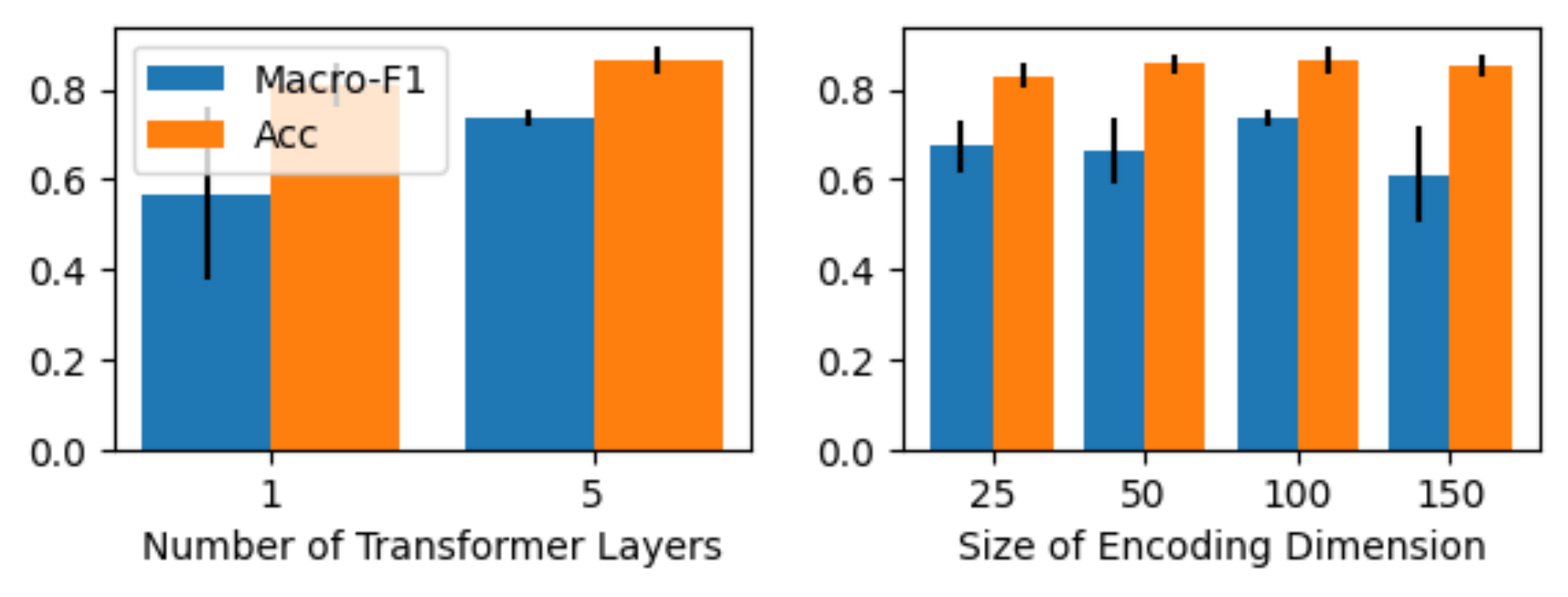}
    
    \caption{Macro-F1 and Accuracy scores for important hyperparameters for \modelshort. (Left) Multi-layered transformers and (Right) encoding dimension of $d_x = 100$ boosts performance.}
    
    \label{fig:hyperparam}
    \vspace{-7mm}
    \end{center}
\end{figure}

\paragraph{Encoding Size \& Transformer Layers} 
\label{subsec:hyperparam}
In Figure~\ref{fig:hyperparam}, we first display results for ablations on varying number transformer encoder layers $M$. The first layer of \cptshorts encodes the $self$'s behavior contingent on $other$. The later layers further contextualizes the contingent $self$'s behavior with its own behavior again, allowing it to attend to other parts of its own behavioral sequence. We find that having multiple layers of transformer encoders leads to a significant improvement in Macro-F1. Secondly, we also display results after varying the size of the embedding dimension per timestep $d_x$, which is an important hyperparameter that controls the expressivity of the model. We find that the optimal encoding dimension to encode behavior per timestep setting is $d_x = 100$.

\vspace{-1mm}
\paragraph{Comparisons against Action Recognition Models}
Following recent approaches in utilizing action recognition models in engagement prediction \cite{ai2022class,kim2022joint}, we compare \modelshorts to activity recognition models in Table \ref{tab:action_reg}. Training state-of-the-art action recognition models are computationally much more expensive than engagement prediction models, due to the fact that the model is trained on a time-series of raw pixels end-to-end. Therefore, instead of applying an 8 seconds window with 1 second interval, we apply an interval of 8 seconds. Even with a modified set-up, there is a large discrepancy between the training time between these two classes of models, one seed takes $\sim150$ minutes for a raw video-based model, compared to $\sim20$ minutes for an engagement prediction model. Nonetheless, for a fair comparison, we train our proposed model with the same training settings. We find that the training of raw video models end-to-end yields poor performance specifically for scarce labels in disengagement.  \modelshorts   performs better than other  architectures given the same training conditions.

% \textbf{encoding size == how expressive it is} \textbf{layers: 1st is crossperson, 2nd onwards is more self-attention like}

% \usepackage{booktabs}
\begin{table}[]
\resizebox{\columnwidth}{!}{%
% \usepackage{booktabs}

\begin{tabular}{@{}l|lll@{}}

\toprule
Model       & Accuracy     & Weighted F1  & Macro F1     \\ \midrule
I3D \cite{NonLocal2018}        & 0.751 $\pm$ 0.07 & 0.658 $\pm$ 0.08 & 0.254 $\pm$ 0.05 \\
TimeSformer \cite{bertasius2021spacetime} & 0.806 $\pm$ 0.03 & 0.752 $\pm$ 0.05 & 0.337 $\pm$ 0.14 \\
SlowFast \cite{feichtenhofer2019slowfast}    & 0.718 $\pm$ 0.11 & 0.628 $\pm$ 0.12 & 0.232 $\pm$ 0.02 \\ \midrule
\modelshorts (Ours)        & \textbf{0.828 $\pm$ 0.02} & \textbf{0.823 $\pm$ 0.02} & \textbf{0.466 $\pm$ 0.06} \\ \bottomrule
\end{tabular}
}
\vspace{-3mm}
\caption{Raw video-based action recognition models and \modelshorts trained with less computationally heavy training set-up. Results and standard deviation are reported for 3 seeds. We see the limitations of training end-to-end raw video-based models.}
\vspace{-4mm}
\label{tab:action_reg}
\end{table}

% \cite{bertasius2021spacetime}, SlowFast \cite{feichtenhofer2019slowfast} and . 

% . , which also require spatio-temporal understanding of human behavior from videos.

\vspace{-2mm}
\subsection{Qualitative Analyses}
Given the upcoming EU regulations \cite{european2021proposal} requiring explainability for any affect-related AI, an important facet of our work is that the resulting attention weights can be used as a way \cite{wiegreffe2019attention} to explain why the model made this specific prediction for this specific timestep.  Given an AI agent which utilizes \modelshorts as its backbone engagement detection module, if a person inquires the agent why it thought that they were disengaged at a point in time, the AI agent could examine is \cpas weights and provide some rationale behind its prediction based on discovered contingent behaviors. 

% based on the behaviors of the target person and other people's behaviors.

% specifically, a high activation from the (1) self-attention weights indicates that certain self-induced behavior mainly drove the model's prediction, and a high activation from the (2) \cpas indicates that contingent behavior from other's behavior mainly drove the predictions.

% (i. e. at timestep X, another person laughed, and the person, in turn, looked at them, which led them to believe that they were disengaged at the task). 

\begin{figure}[t]
    \vspace{-0mm}
    \begin{center}
    \includegraphics[width=1.1\columnwidth]{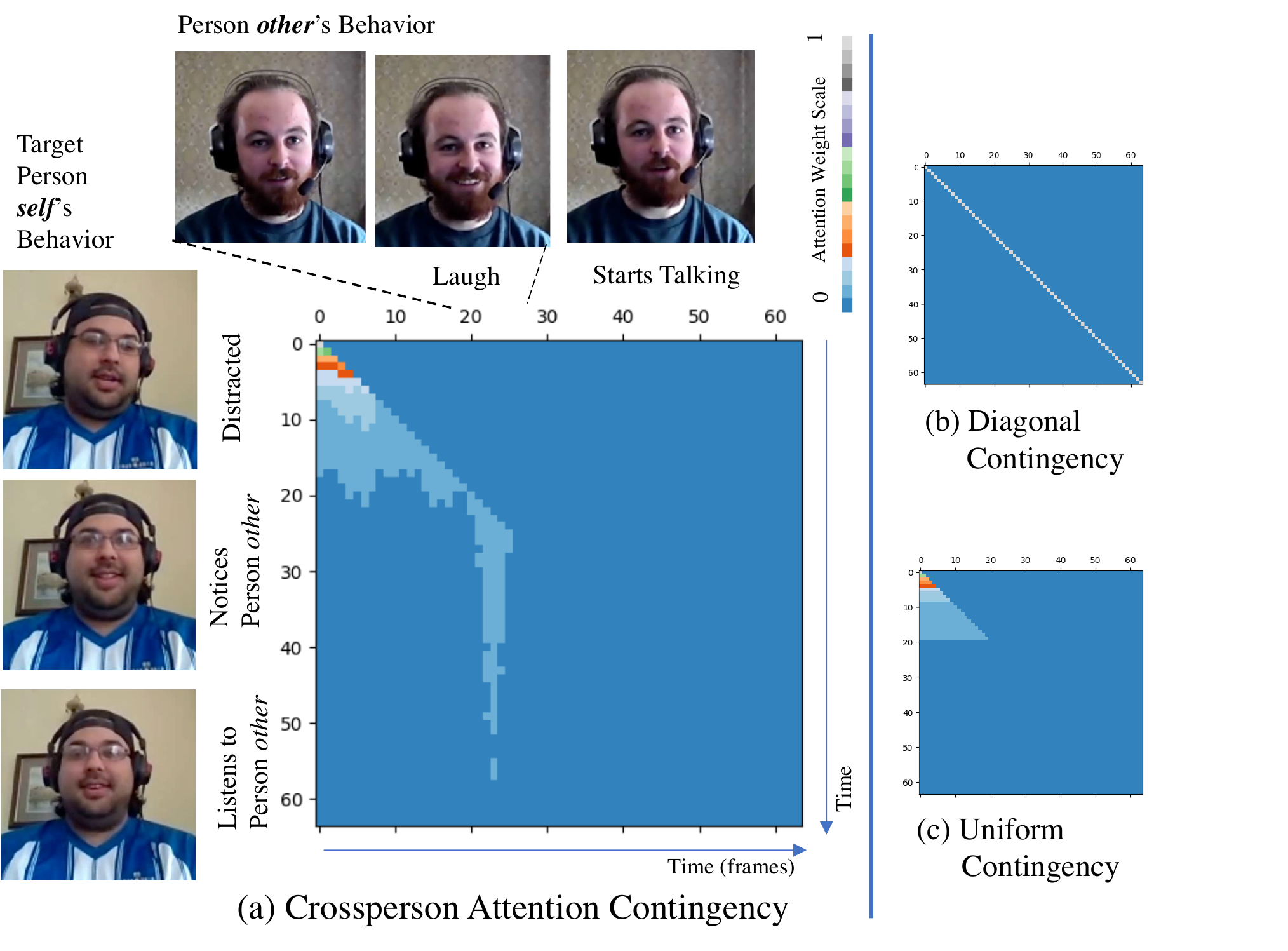}
    \vspace{-6mm}
    \caption{(a) \modelss Cross-person Attention weights from $t=120s$ for group $S01$. \modelshorts has discovered that $self$’s behavior (smile and listen) from timestep 16 – 60 is contingent on $other$’s behavior in timestep 20 – 25 (laughter). (b) Diagonal Contingency: Crossperson attention weights with the assumption that person $self$ and $other$’s behavior are only related at the exact same timesteps. (c) Uniform Contingency: Default behavior of Crossperson attention weights; i.e. all of person $self$’s past behaviors are uniformly related to person $other$’s current behavior.
}
    \vspace{-6mm}
    \label{fig:contingency}
    \end{center}
\end{figure}

To demonstrate how \cpas (CPA) could be used to explain model predictions, in Figure~\ref{fig:contingency}, we visualize the attention weights from the first layer of the \cpas between Person $self$, (left) and Person $other$, (top) and show that it is able to capture contingent behavior between pairs of individuals' behaviors. The $x$ and $y$ axes refer to each person's behavior aligned to timesteps. We first provide attention weights that demonstrate the lack of contingency for comparison. The attention weights visualized in Figure~\ref{fig:contingency}(b) Diagonal Contingency, is a diagonal attention weight matrix, which indicates that Person $self$  and Person $other$'s behavior are only related at the exact same time steps. The attention weights visualized in Figure~\ref{fig:contingency}(c), Uniform Contingency, is the default behavior if we assume that all of Person $other$’s past behavior is uniformly related to Person $self$’s current behavior, demonstrated by the uniform color across each row, which indicates that the attention weights are uniformly distributed across the available timesteps. The upper triangular matrix is masked to encode the natural assumption that $other$'s future behavior shouldn't affect $self$'s current behavior.

Figure~\ref{fig:contingency}(a) shows the learnt \cpas weights, which indicates that Person $self$’s behavior in timesteps $16–60$ is contingent on Person $other$’s behavior in timestep $20-25$. We find that, after inspecting the video at the aligned timesteps, that Person $other$ laughs during timestep $20-25$ and starts to talk afterwards. Person $self$ was initially distracted, but after they see Person $other$ laughing at timestep $20-25$, they look at Person $other$ and starts listening. Hence, we find that \cpas has discovered meaningful contingent behavior between two people. We kindly refer the reader to the supplementary for more examples.

% \paragraph{Explainability} 

% \modelshorts allows 
% which I think is quite valuable, especially in Human-AI interaction settings. . I now think inclusion of this discussion would be valuable for the paper-- thank you! What were your other thoughts? 

% When predicting low engagement, 

% the model that can attend to the $self$'s influence on other's behaviors results in higher performance. 

% All of our models 

% - engagement prediction
% - in depth class by class labels

% - video-based 
% - see limitations, macro-f1 unfortunately lower 

% \subsection{Direction of Cross-person Attention}
% CPA_{other \rightarrow self }: Query: other , key: self, value: self CPA_{self \leftarrow other }

%  We replace the directionality of crossperson attention to see if

% how I affect others is more important or how others affect me are more important in engagement prediction. No question, easily how others affect me should be more important. but what is fascinating is that in the case for predicting disengagement, how I affect others is more important as indicated by the significant drop in dis engagement f1  scores. On the other hand, for engaged instances, how other people impact me is more important 

% \section{Discussion}

\vspace{-2mm}
\section{Conclusion}
\vspace{-1mm}

In this work, we study the challenging task of modelling human behaviors in a multiparty setting. We proposed a new Transformer-based model for multiparty behavior modeling, \modelshort. At the core of \modelshorts is \cpa, which is designed to capture contingent behaviors. We compare \modelshorts against previous approaches on a timely and challenging task of engagement prediction in online meetings and provide in-depth analysis and ablation studies. We see significant gains in performance (up to 10\%) compared to previous approaches, where we find that controlling the direction of \cpa, including multiple layers of transformer blocks, and self attention blocks is crucial. We also demonstrate qualitatively that our model is able to find contingent behaviors. Our \modelshorts is a novel approach to modeling contingent behaviors in multiparty conversation, a crucial problem in developing AI agents that can communicate with groups of people. We publicly share our code to enable research in multiparty interactions.

\paragraph{Limitations and Future Work} Our evaluation benchmark Roomreader \cite{reverdy2022roomreader} is collected from online lessons, which is one specific context. Future works should test the generalizability of models in different situations and scenarios including in-person settings, various relationships, diverse cultures, and a wider range of age of participants. We believe that our proposed method would generalize across different group settings, environment, and tasks, as supported by literature reviewed in Section~\ref{sec:related_contingent_behavior} that contingent behaviors play an important role in various interaction settings. 
%in groups of peers \cite{sage1999peer}, teachers and students \cite{boyd2006contingent},  caregivers and infants \cite{masek2021language}, children and on-screen characters \cite{xu2022contingent}. 
In addition, here we present results on the 5-person setting, as the dataset offered the most amount of data for this specific size. Follow-up work should focus on developing approaches that can perform well with a variable number of people. Using a unified encoder for all individual behavior as well as a unified cross-person transformer for every pairwise behavior is promising. This would alleviate the need to train an $N$-person-specific model. In this work, we purely rely on the visual modality. Contingent behavior exists in language and acoustics as well. Using a multimodal input which includes these modalities should be investigated.

% Future work should also investigate how these contingencies and behavioral patterns differ. 

% \paragraph{Subjectivity of Engagement} As mentioned in Section \ref{sec:related}, engagement is a multi-dimensional construct that can be measured in various ways. Furthermore, the target towards which a person is engaged is crucial. In our work, we study on-task, off-task engagement as defined by the RoomReader \cite{reverdy2022roomreader} dataset. Future work should delve into different types of engagement, as they could require different types of reasoning or inductive biases.

% \section{Conclusion }

\bibliographystyle{named}
\bibliography{ijcai23}

\begin{thebibliography}{}

\bibitem[\protect\citeauthoryear{Adams \bgroup \em et al.\egroup
  }{2006}]{Adamssmallgroup}
Katherine~L Adams, Gloria~J Galanes, and John~K Brilhart.
\newblock {\em Communicating in groups: Applications and skills. Chapter 4:
  Using Verbal and Nonverbal Messages in a Group.}
\newblock McGraw-Hill Boston, 2006.

\bibitem[\protect\citeauthoryear{Admoni and Scassellati}{2014}]{admoni2014data}
Henny Admoni and Brian Scassellati.
\newblock Data-driven model of nonverbal behavior for socially assistive
  human-robot interactions.
\newblock In {\em Proc. of the 16th int. conf. on multimodal interaction
  (ICMI)}, 2014.

\bibitem[\protect\citeauthoryear{Ai \bgroup \em et al.\egroup
  }{2022}]{ai2022class}
Xusheng Ai, Victor~S Sheng, and Chunhua Li.
\newblock Class-attention video transformer for engagement intensity
  prediction.
\newblock {\em arXiv:2208.07216}, 2022.

\bibitem[\protect\citeauthoryear{Baltrusaitis \bgroup \em et al.\egroup
  }{2018}]{8373812}
Tadas Baltrusaitis, Amir Zadeh, Yao~Chong Lim, and Louis-Philippe Morency.
\newblock Openface 2.0: Facial behavior analysis toolkit.
\newblock In {\em Proc. of the 13th IEEE International Conference on Automatic
  Face and Gesture Recognition (FG)}, pages 59--66, 2018.

\bibitem[\protect\citeauthoryear{Bernieri \bgroup \em et al.\egroup
  }{1994}]{bernieri1994interactional}
Frank~J Bernieri, Janet~M Davis, Robert Rosenthal, and C~Raymond Knee.
\newblock Interactional synchrony and rapport: Measuring synchrony in displays
  devoid of sound and facial affect.
\newblock {\em Personality and social psychology bulletin}, 20(3):303--311,
  1994.

\bibitem[\protect\citeauthoryear{Bernieri}{1988}]{bernieri1988coordinated}
Frank~J Bernieri.
\newblock {\em Coordinated movement in human interaction: Synchrony, posture
  similarity, and rapport}.
\newblock Harvard University, 1988.

\bibitem[\protect\citeauthoryear{Bertasius \bgroup \em et al.\egroup
  }{2021}]{bertasius2021spacetime}
Gedas Bertasius, Heng Wang, and Lorenzo Torresani.
\newblock Is space-time attention all you need for video understanding?, 2021.

\bibitem[\protect\citeauthoryear{Boyd and Rubin}{2006}]{boyd2006contingent}
Maureen Boyd and Don Rubin.
\newblock How contingent questioning promotes extended student talk: A function
  of display questions.
\newblock {\em Journal of Literacy Research}, 38(2):141--169, 2006.

\bibitem[\protect\citeauthoryear{Chapman}{1997}]{chapman1997models}
Peter~McFaul Chapman.
\newblock {\em Models of engagement: Intrinsically motivated interaction with
  multimedia learning software}.
\newblock PhD thesis, University of Waterloo, 1997.

\bibitem[\protect\citeauthoryear{Chen \bgroup \em et al.\egroup
  }{2020a}]{chen2020transformer}
Haifeng Chen, Dongmei Jiang, and Hichem Sahli.
\newblock Transformer encoder with multi-modal multi-head attention for
  continuous affect recognition.
\newblock {\em IEEE Transactions on Multimedia}, 23:4171--4183, 2020.

\bibitem[\protect\citeauthoryear{Chen \bgroup \em et al.\egroup
  }{2020b}]{chen2020teaching}
Huili Chen, Hae~Won Park, and Cynthia Breazeal.
\newblock Teaching and learning with children: Impact of reciprocal peer
  learning with a social robot on children’s learning and emotive engagement.
\newblock {\em Computers \& Education}, 150:103836, 2020.

\bibitem[\protect\citeauthoryear{Chen \bgroup \em et al.\egroup
  }{2021}]{chen2021crossvit}
Chun-Fu~Richard Chen, Quanfu Fan, and Rameswar Panda.
\newblock Crossvit: Cross-attention multi-scale vision transformer for image
  classification.
\newblock In {\em Proceedings of the IEEE/CVF international conference on
  computer vision}, pages 357--366, 2021.

\bibitem[\protect\citeauthoryear{Chen \bgroup \em et al.\egroup
  }{2022}]{chen2022dyadic}
Huili Chen, Sharifa~Mohammed Alghowinem, Soo~Jung Jang, Cynthia Breazeal, and
  Hae~Won Park.
\newblock Dyadic affect in parent-child multi-modal interaction: Introducing
  the dami-p2c dataset and its preliminary analysis.
\newblock {\em IEEE Transactions on Affective Computing}, 2022.

\bibitem[\protect\citeauthoryear{Corno and Mandinach}{1983}]{corno1983role}
Lyn Corno and Ellen~B Mandinach.
\newblock The role of cognitive engagement in classroom learning and
  motivation.
\newblock {\em Educational psychologist}, 18(2):88--108, 1983.

\bibitem[\protect\citeauthoryear{Curto \bgroup \em et al.\egroup
  }{2021}]{curto2021dyadformer}
David Curto, Albert Clap{\'e}s, Javier Selva, Sorina Smeureanu, Julio Junior,
  CS~Jacques, David Gallardo-Pujol, Georgina Guilera, David Leiva, Thomas~B
  Moeslund, et~al.
\newblock Dyadformer: A multi-modal transformer for long-range modeling of
  dyadic interactions.
\newblock In {\em Proceedings of the IEEE/CVF International Conference on
  Computer Vision}, pages 2177--2188, 2021.

\bibitem[\protect\citeauthoryear{Del~Duchetto \bgroup \em et al.\egroup
  }{2020}]{del2020you}
Francesco Del~Duchetto, Paul Baxter, and Marc Hanheide.
\newblock Are you still with me? continuous engagement assessment from a
  robot's point of view.
\newblock {\em Frontiers in Robotics and AI}, 7:116, 2020.

\bibitem[\protect\citeauthoryear{Dhall \bgroup \em et al.\egroup
  }{2020}]{dhall2020emotiw}
Abhinav Dhall, Garima Sharma, Roland Goecke, and Tom Gedeon.
\newblock Emotiw 2020: Driver gaze, group emotion, student engagement and
  physiological signal based challenges.
\newblock In {\em Proceedings of the 2020 International Conference on
  Multimodal Interaction}, 2020.

\bibitem[\protect\citeauthoryear{EUCommission}{2021}]{european2021proposal}
EUCommission.
\newblock {Proposal for a Regulation of the European Parliament and of the
  Council, Laying Down Harmonised Rules on Artificial Intelligence (Artificial
  Intelligence Act) and Amending Certain Union Legislative Acts, SEC (2021) 167
  final, COM (2021) 2006 final}, 2021.

\bibitem[\protect\citeauthoryear{Feichtenhofer \bgroup \em et al.\egroup
  }{2019}]{feichtenhofer2019slowfast}
Christoph Feichtenhofer, Haoqi Fan, Jitendra Malik, and Kaiming He.
\newblock Slowfast networks for video recognition.
\newblock In {\em Proceedings of the IEEE international conference on computer
  vision}, 2019.

\bibitem[\protect\citeauthoryear{Fredricks \bgroup \em et al.\egroup
  }{2004}]{fredricks2004school}
Jennifer~A Fredricks, Phyllis~C Blumenfeld, and Alison~H Paris.
\newblock School engagement: Potential of the concept, state of the evidence.
\newblock {\em Review of educational research}, 74(1):59--109, 2004.

\bibitem[\protect\citeauthoryear{Griffin \bgroup \em et al.\egroup
  }{2008}]{griffin2008behavioral}
Mark~A Griffin, Sharon~K Parker, and Andrew Neal.
\newblock Is behavioral engagement a distinct and useful construct?
\newblock {\em Industrial and Organizational Psychology}, 1(1):48--51, 2008.

\bibitem[\protect\citeauthoryear{He \bgroup \em et al.\egroup
  }{2016}]{he2016deep}
Kaiming He, Xiangyu Zhang, Shaoqing Ren, and Jian Sun.
\newblock Deep residual learning for image recognition.
\newblock In {\em Proceedings of the IEEE conference on computer vision and
  pattern recognition}, pages 770--778, 2016.

\bibitem[\protect\citeauthoryear{Kim \bgroup \em et al.\egroup
  }{2022}]{kim2022joint}
Yubin Kim, Huili Chen, Sharifa Alghowinem, Cynthia Breazeal, and Hae~Won Park.
\newblock Joint engagement classification using video augmentation techniques
  for multi-person human-robot interaction.
\newblock {\em arXiv:2212.14128}, 2022.

\bibitem[\protect\citeauthoryear{Kopp}{2010}]{kopp2010social}
Stefan Kopp.
\newblock Social resonance and embodied coordination in face-to-face
  conversation with artificial interlocutors.
\newblock {\em Speech Communication}, 52(6):587--597, 2010.

\bibitem[\protect\citeauthoryear{Lakin \bgroup \em et al.\egroup
  }{2003}]{lakin2003chameleon}
Jessica~L Lakin, Valerie~E Jefferis, Clara~Michelle Cheng, and Tanya~L
  Chartrand.
\newblock The chameleon effect as social glue: Evidence for the evolutionary
  significance of nonconscious mimicry.
\newblock {\em Journal of nonverbal behavior}, 27(3):145--162, 2003.

\bibitem[\protect\citeauthoryear{Li \bgroup \em et al.\egroup }{2020}]{gr_sr1}
Wanhua Li, Yueqi Duan, Jiwen Lu, Jianjiang Feng, and Jie Zhou.
\newblock Graph-based social relation reasoning.
\newblock In {\em European Conference on Computer Vision}, pages 18--34.
  Springer, 2020.

\bibitem[\protect\citeauthoryear{Lin and Lee}{2020}]{gr_perf}
Yun-Shao Lin and Chi-Chun Lee.
\newblock Predicting performance outcome with a conversational graph
  convolutional network for small group interactions.
\newblock In {\em 2020 IEEE International Conference on Acoustics, Speech and
  Signal Processing (ICASSP)}, 2020.

\bibitem[\protect\citeauthoryear{Lin \bgroup \em et al.\egroup
  }{2017}]{lin2017focal}
Tsung-Yi Lin, Priya Goyal, Ross Girshick, Kaiming He, and Piotr Doll{\'a}r.
\newblock Focal loss for dense object detection.
\newblock In {\em Proceedings of the IEEE international conference on computer
  vision}, pages 2980--2988, 2017.

\bibitem[\protect\citeauthoryear{Loshchilov and
  Hutter}{2017}]{loshchilov2017decoupled}
Ilya Loshchilov and Frank Hutter.
\newblock Decoupled weight decay regularization.
\newblock {\em arXiv preprint arXiv:1711.05101}, 2017.

\bibitem[\protect\citeauthoryear{Ma \bgroup \em et al.\egroup
  }{2021}]{ma2021hierarchical}
Jiayao Ma, Xinbo Jiang, Songhua Xu, and Xueying Qin.
\newblock Hierarchical temporal multi-instance learning for video-based student
  learning engagement assessment.
\newblock In {\em IJCAI}, pages 2782--2789, 2021.

\bibitem[\protect\citeauthoryear{Masek \bgroup \em et al.\egroup
  }{2021}]{masek2021language}
Lillian~R Masek, Brianna~TM McMillan, Sarah~J Paterson, Catherine~S
  Tamis-LeMonda, Roberta~Michnick Golinkoff, and Kathy Hirsh-Pasek.
\newblock Where language meets attention: How contingent interactions promote
  learning.
\newblock {\em Developmental Review}, 60:100961, 2021.

\bibitem[\protect\citeauthoryear{Park \bgroup \em et al.\egroup
  }{2012}]{park2012makes}
Sira Park, Susan~D Holloway, Amanda Arendtsz, Janine Bempechat, and Jin Li.
\newblock What makes students engaged in learning? a time-use study of
  within-and between-individual predictors of emotional engagement in
  low-performing high schools.
\newblock {\em Journal of youth and adolescence}, 41(3):390--401, 2012.

\bibitem[\protect\citeauthoryear{Park \bgroup \em et al.\egroup
  }{2017}]{park2017hri}
Hae~Won Park, Mirko Gelsomini, Jin~Joo Lee, and Cynthia Breazeal.
\newblock Telling stories to robots: The effect of backchanneling on a child's
  storytelling.
\newblock In {\em Proc. of the ACM/IEEE Int. Conf. on Human-Robot Interaction
  (HRI)}, 2017.

\bibitem[\protect\citeauthoryear{Ren}{2016}]{ren2016rethinking}
Xiangshi Ren.
\newblock Rethinking the relationship between humans and computers.
\newblock {\em Computer}, 49(8):104--108, 2016.

\bibitem[\protect\citeauthoryear{Reverdy \bgroup \em et al.\egroup
  }{2022}]{reverdy2022roomreader}
Justine Reverdy, Sam~O’Connor Russell, Louise Duquenne, Diego Garaialde,
  Benjamin~R Cowan, and Naomi Harte.
\newblock Roomreader: A multimodal corpus of online multiparty conversational
  interactions.
\newblock In {\em Proceedings of the Thirteenth Language Resources and
  Evaluation Conference}, pages 2517--2527, 2022.

\bibitem[\protect\citeauthoryear{Sage and Kindermann}{1999}]{sage1999peer}
Nicole~A Sage and Thomas~A Kindermann.
\newblock Peer networks, behavior contingencies, and children's engagement in
  the classroom.
\newblock {\em Merrill-Palmer Quarterly (1982-)}, pages 143--171, 1999.

\bibitem[\protect\citeauthoryear{Silpasuwanchai \bgroup \em et al.\egroup
  }{2016}]{silpasuwanchai2016developing}
Chaklam Silpasuwanchai, Xiaojuan Ma, Hiroaki Shigemasu, and Xiangshi Ren.
\newblock Developing a comprehensive engagement framework of gamification for
  reflective learning.
\newblock In {\em Proc. of the ACM Conf. on Designing Interactive Systems},
  pages 459--472, 2016.

\bibitem[\protect\citeauthoryear{Steinert \bgroup \em et al.\egroup
  }{2020}]{steinert2020towards}
Lars Steinert, Felix Putze, Dennis K{\"u}ster, and Tanja Schultz.
\newblock Towards engagement recognition of people with dementia in care
  settings.
\newblock In {\em Proceedings of the 2020 International Conference on
  Multimodal Interaction}, pages 558--565, 2020.

\bibitem[\protect\citeauthoryear{Thong~Huynh \bgroup \em et al.\egroup
  }{2019}]{thong2019engagement}
Van Thong~Huynh, Soo-Hyung Kim, Guee-Sang Lee, and Hyung-Jeong Yang.
\newblock Engagement intensity prediction withfacial behavior features.
\newblock In {\em 2019 International Conference on Multimodal Interaction},
  pages 567--571, 2019.

\bibitem[\protect\citeauthoryear{Tsai \bgroup \em et al.\egroup
  }{2019}]{tsai2019multimodal}
Yao-Hung~Hubert Tsai, Shaojie Bai, Paul~Pu Liang, J~Zico Kolter, Louis-Philippe
  Morency, and Ruslan Salakhutdinov.
\newblock Multimodal transformer for unaligned multimodal language sequences.
\newblock In {\em Proceedings of the conference. Association for Computational
  Linguistics. Meeting}, volume 2019, page 6558, 2019.

\bibitem[\protect\citeauthoryear{Vaswani \bgroup \em et al.\egroup
  }{2017}]{vaswani2017attention}
Ashish Vaswani, Noam Shazeer, Niki Parmar, Jakob Uszkoreit, Llion Jones,
  Aidan~N Gomez, {\L}ukasz Kaiser, and Illia Polosukhin.
\newblock Attention is all you need.
\newblock {\em Advances in neural information processing systems}, 30, 2017.

\bibitem[\protect\citeauthoryear{Wang \bgroup \em et al.\egroup
  }{2018}]{NonLocal2018}
Xiaolong Wang, Ross Girshick, Abhinav Gupta, and Kaiming He.
\newblock Non-local neural networks.
\newblock {\em CVPR}, 2018.

\bibitem[\protect\citeauthoryear{Wang \bgroup \em et al.\egroup
  }{2019}]{wang2019bootstrap}
Kai Wang, Jianfei Yang, Da~Guo, Kaipeng Zhang, Xiaojiang Peng, and Yu~Qiao.
\newblock Bootstrap model ensemble and rank loss for engagement intensity
  regression.
\newblock In {\em 2019 International Conference on Multimodal Interaction},
  pages 551--556, 2019.

\bibitem[\protect\citeauthoryear{Wiegreffe and
  Pinter}{2019}]{wiegreffe2019attention}
Sarah Wiegreffe and Yuval Pinter.
\newblock Attention is not not explanation.
\newblock {\em arXiv preprint arXiv:1908.04626}, 2019.

\bibitem[\protect\citeauthoryear{Xu \bgroup \em et al.\egroup
  }{2022}]{xu2022contingent}
Ying Xu, Valery Vigil, Andres~S Bustamante, and Mark Warschauer.
\newblock Contingent interaction with a television character promotes
  children's science learning and engagement.
\newblock {\em Journal of Applied Developmental Psychology}, 81:101439, 2022.

\bibitem[\protect\citeauthoryear{Yang \bgroup \em et al.\egroup
  }{2020}]{gr_behav}
Fangkai Yang, Wenjie Yin, Tetsunari Inamura, M{\aa}rten Bj{\"o}rkman, and
  Christopher Peters.
\newblock Group behavior recognition using attention-and graph-based neural
  networks.
\newblock In {\em ECAI 2020}, pages 1626--1633. 2020.

\bibitem[\protect\citeauthoryear{Zhang \bgroup \em et al.\egroup
  }{2022}]{zhang2022engagement}
Zhijie Zhang, Jianmin Zheng, and Nadia Magnenat~Thalmann.
\newblock Engagement estimation of the elderly from wild multiparty
  human--robot interaction.
\newblock {\em Computer Animation and Virtual Worlds}, page e2120, 2022.

\bibitem[\protect\citeauthoryear{Zhou \bgroup \em et al.\egroup
  }{2019}]{zhou2019social}
Chen Zhou, Ming Han, Qi~Liang, Yi-Fei Hu, and Shu-Guang Kuai.
\newblock A social interaction field model accurately identifies static and
  dynamic social groupings.
\newblock {\em Nature human behaviour}, 3(8):847--855, 2019.

\end{thebibliography}

% \appendix

% \input{99_appendix.tex}

\end{document}